\documentclass[letterpaper, 10 pt, conference]{ieeeconf}  

\IEEEoverridecommandlockouts                              

\usepackage{graphics} 
\usepackage{epsfig} 
\usepackage{mathptmx} 
\usepackage{times} 
\usepackage{amsmath} 
\usepackage{amssymb}  
\usepackage[colorlinks,linkcolor=blue]{hyperref} 
\usepackage{utfsym}
\usepackage{times} 

\title{\LARGE \bf
LiteFat: Lightweight Spatio-Temporal Graph Learning for Real-Time Driver Fatigue Detection
}

\author{Jing Ren$^{1}$, Suyu Ma$^{2}$, Hong Jia$^{3}$, Xiwei Xu$^{2}$, Ivan Lee$^{4}$, Haytham Fayek$^{1}$, Xiaodong Li$^{1}$, and Feng Xia$^{1}$
\thanks{$^{1}$Jing Ren, Haytham Fayek, Xiaodong Li, and Feng Xia are with School of Computing Technologies,
RMIT University, Australia.
        {\tt\small jing.ren@ieee.org, haytham.fayek@ieee.org, xiaodong.li@rmit.edu.au, f.xia@ieee.org}}%
\thanks{$^{2}$Suyu Ma and Xiwei Xu are with CSIRO's Data61, Australia.
        {\tt\small suyu.ma@data61.csiro.au, xiwei.xu@data61.csiro.au}}%
\thanks{$^{3}$Hong Jia is with the School of Computer Science, University of Auckland, New Zealand.
        {\tt\small hong.jia@auckland.ac.nz}}%
  \thanks{$^{4}$ Ivan Lee is with STEM, University of South Australia, Australia.
        {\tt\small ivan.lee@unisa.edu.au}}%
}

\begin{document}

\maketitle
\thispagestyle{empty}
\pagestyle{empty}

\begin{abstract}
Detecting driver fatigue is critical for road safety, as drowsy driving remains a leading cause of traffic accidents. Many existing solutions rely on computationally demanding deep learning models, which result in high latency and are unsuitable for embedded robotic devices with limited resources (such as intelligent vehicles/cars) where rapid detection is necessary to prevent accidents. This paper introduces LiteFat, a lightweight spatio-temporal graph learning model designed to detect driver fatigue efficiently while maintaining high accuracy and low computational demands. LiteFat involves converting streaming video data into spatio-temporal graphs (STG) using facial landmark detection, which focuses on key motion patterns and reduces unnecessary data processing. LiteFat uses MobileNet to extract facial features and create a feature matrix for the STG. A lightweight spatio-temporal graph neural network is then employed to identify signs of fatigue with minimal processing and low latency. Experimental results on benchmark datasets show that LiteFat performs competitively while significantly reduced computational complexity and latency as compared to current state-of-the-art methods. This work advances the development of real-time, resource-efficient human fatigue detection systems that can be implemented upon embedded robotic devices.

\end{abstract}

\section{Introduction}


Driver fatigue is a major contributor to traffic accidents worldwide, posing a significant threat to road safety~\cite{fu2024survey}. The World Health Organization (WHO) reports that traffic accidents result in 1.35 million fatalities each year globally\footnote{\href{https://www.who.int/activities/advocating-for-road-safety}{https://www.who.int/activities/advocating-for-road-safety}}. Research further highlights that fatigue-related incidents can account for up to 30\% of these fatal accidents~\cite{nvemcova2020multimodal}. Fatigue impairs reaction time, decision-making, and overall driving performance, underscoring the necessity for real-time fatigue detection to prevent such accidents. Even with advancements in autonomous driving technology, human oversight is still crucial, particularly in semi-autonomous vehicles.


In recent years, artificial intelligence (AI)-based human fatigue detection systems have been developed, leveraging both behavioral characteristics~\cite{liu2022review} and physiological signals~\cite{lu2022detecting,hu2024eeg,peng2024multi}. Physiological methods, such as electroencephalography (EEG) and heart rate monitoring~\cite{subasi2022eeg}, offer high accuracy but require intrusive sensors, making them impractical for real-world deployment. On the other hand, methods based on behavioral characteristics usually rely on cameras to capture driver's videos, which are more suitable for real-world applications. Vision-based models analyze facial features, such as eye closure, yawning, and head position, using deep learning techniques and have gained popularity~\cite{ansari2021driver,ren2021deep}. When integrated into early warning systems, these models can monitor a driver's state in real time and issue timely warnings. This automation process holds significant practical value in reducing traffic accidents and improving road safety. 


Despite significant advancements in vision-based fatigue detection, numerous practical challenges remain \cite{sikander2018driver,zhang2022systematic,kaur2024deepfake}. These models often require substantial computational power to process high-resolution video data in real time, leading to high latency and memory demands, which complicates deployment on embedded automotive systems or robotic devices \cite{shahzad2024decade,li2024gmmap,yamada2024twist}. To enhance efficiency, some approaches convert video data into spatio-temporal graphs (STG) and employ graph learning models to capture spatial and temporal relationships \cite{ren2023graph}. However, these methods often use predefined spatial links for facial landmarks, introducing human bias~\cite{xu2025fairness} and potentially missing critical information for fatigue detection. Additionally, real-world challenges such as variations in head poses, facial expressions, lighting conditions, and occlusions from glasses or masks can significantly hinder the accuracy and reliability of vision-based driver fatigue detection systems, making it an ongoing issue in the field~\cite{lin2025early}.


To address these limitations, this paper introduces LiteFat, a lightweight AI model designed for efficient driver fatigue detection in embedded robotic systems. The model consists of three main modules (see Fig.~\ref{fig:fm}). First, the \textit{key frame selection and facial landmark detection module} extracts key frames from each video clip to enhance efficiency. Second, the \textit{multimodal feature fusion for feature matrix construction module} generates an informative feature matrix for the facial graphs in each frame. Finally, the \textit{spatio-temporal graph learning for fatigue classification module} classifies driver behaviors into categories including normal, yawning, and talking. Unlike existing models, this module adaptively learns the adjacency matrix of each frame during the learning process, rather than relying on predefined configurations. LiteFat is designed to operate in real-time on embedded robotic devices, achieving a balance between accuracy and computational efficiency. By integrating MobileNet with spatio-temporal graph learning (STGL) models, the approach minimizes processing power requirements while maintaining reliable fatigue detection performance, setting it apart from conventional deep learning models that typically demand extensive computational resources.

The main contributions of this paper are as follows:
\begin{itemize}
    \item We propose a lightweight graph learning human fatigue detection model called LiteFat, which can achieve high accuracy while maintaining low computational complexity, making it suitable for real-time deployment on embedded robotic devices. 
\item We introduce a multimodal fusion approach that dynamically constructs feature matrices from facial landmarks and MobileNet-based facial embeddings. Instead of using predefined facial graph structures, LiteFat learns the adjacency matrix adaptively during training, reducing human bias and capturing more accurate spatial relationships.

    \item We conduct comprehensive experiments on a benchmark driver fatigue dataset. LiteFat demonstrates state-of-the-art (SOTA) performance, while significantly improving computational efficiency and reducing latency compared to baseline methods.

\end{itemize}

The remainder of this paper is structured as follows. Section~\ref{sec2} brieﬂy reviews related work. The constructions of feature matrix and the spatio-temporal graph learning model for fatigue detection are described in detail in Section~\ref{sec3}. In Section~\ref{sec4}, the experimental details and results are reported. Section~\ref{sec5} summarizes the paper. 

\section{Related Work}~\label{sec2}
Driver behavior detection has been widely studied using various techniques~\cite{wang2021survey}. Although deep learning models have shown promising results in accuracy, they often suffer from high computational costs, making them unsuitable for real-time applications in embedded robotic systems. This section reviews recent advancements in vision-based driver fatigue detection and lightweight object detection, discussing their models and limitations, particularly in terms of efficiency, accuracy, and deployment feasibility.
\subsection{Vision-based Driver Fatigue Detection}
Vision-based methods for driver fatigue detection analyze visual cues from facial expressions, eye movements, and head posture to assess the driver's alertness. These approaches leverage computer vision and deep learning techniques to provide non-invasive and real-time monitoring. To effectively capture essential facial features associated with drowsiness while eliminating the influence of irrelevant information, facial landmark detection is applied to convert streaming videos into spatio-temporal graphs before inputting the data to the  models~\cite{zhang2022systematic}. For example, 
Bai et al. \cite{bai2021two} feeded spatial graphs into spatial graph convolutional network (GCN) while feeding the temporal graphs into temporal GCN. This process learns spatial and temporal features separately and uses a softmax score to detect driver drowsiness behaviors. Similarly, 
the approach by Yang et al. \cite{yang2024video} also contains two branches in which the spatial branch encodes the spatial features of facial landmarks. The difference is that the temporal branch process a combined embedding of the facial landmarks and local areas of eyes and mouths instead of temporal graph only.
Lu et al. \cite{lu2023jhpfa} selected the key frames from the original frame sequence by deleting frames similar to adjacent frames. The similarity was evaluated on the basis of the spatial graphs of two consecutive frames. In
\cite{huang2024self}, the authors gave a novel definition of peak frame and non-peak frame, where peak frame contains behavior such as closing eyes and laughing, while non-peak frames only contain general states. Then, they proposed a multi-head graph attention module to distinguish peak frames from non-peak frames  by comparing the spatial graphs.

Apart from GCN, other types of models are also used to learn spatial and temporal features in videos. For example,
Mou et al. \cite{mou2021isotropic} proposed an attention-based multimodal fusion model, where images of eyes, mouths, and optical flow of the driver head are defined as three modalities. In this model, spatial and temporal features are encoded by the convolution neural network (CNN) and long short-term memory (LSTM) networks, respectively. Then the attention layer is linked after LSTM to weigh the contributions of hidden states.
Phan et al. \cite{phan2023driver} designed and integrated an emotion detection module into the system to remove frames showing emotions that are not relevant to drowsiness. This process can shorten the processing time for feature extraction and detection.
 To solve the efficiency problem, Ahmed et al.
\cite{ahmed2021intelligent} focused only on the regions of drivers' eyes and mouths, where both features are learned separately with CNN, and then stacking is employed as the ensemble learning technique. Although this paper considers the issue of model efficiency, focusing only on the driver's mouth and eye features may lead to the loss of key information, which could affect the model's detection performance consequently.
  \begin{figure*}[thpb]
      \centering
      \includegraphics[width=0.96\textwidth]{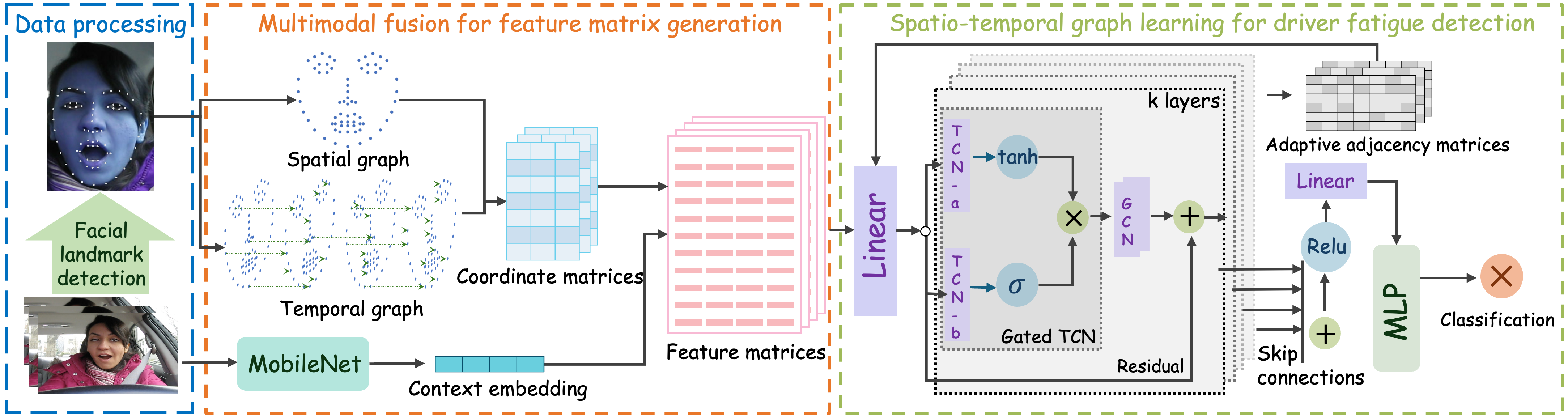}
      \caption{Framework of LiteFat. The first step is facial landmark detection of drivers from streaming video, which will be constructed as spatio-temporal graph for each frame. The second step is to apply MobileNet~\cite{sandler2018mobilenetv2} to capture facial features. The third step is to learn a multimodal fusion module for graph feature matrix construction by combining the coordinates of facial landmarks with features embeddings learned by MobileNet. Then, the spatial and temporal features of nodes will be learned by GraphWavenet for final anomaly score calculation in the last step.}
      \label{fig:fm}
   \end{figure*}
   
\subsection{Spatial Temporal Graph Learning for Human Action Recognition}
Human action recognition (HAR) is a crucial task in, for example, robotics, intelligent vehicles, surveillance, healthcare, and human-computer interaction. To effectively model spatial and temporal dependencies of human motion, STGL leverages GNNs to represent the human body as a graph structure, where joints are nodes and bones are edges. This method effectively captures both spatial relationships and temporal dynamics, which therefore has gained significant popularity in recent years~\cite{feng2022comparative}. In \cite{cheng2020skeleton}, the authors proposed lightweight spatial and temporal shift graph operations to the GCN-based action recognition models with the aim of reducing computational complexity. Similarly,
Zhang et al. \cite{zhang2020semantics} exploited the semantics and dynamics to improve the efficiency of skeleton-based action recognition as well.
Liu et al. \cite{liu2020disentangling} proposed a multi-scale aggregation scheme to avoid biased weighting through removing redundant dependencies.
Chen et al. \cite{chen2021channel} proposed a model to learn a shared topology for the STG with channel-specific correlations. This method avoids modeling each channel independently, which can reduce the parameters and improve the representation capability of GCN model.
Chi et al. \cite{chi2022infogcn} proposed to dynamically infer the intrinsic topology based on context for spatial modeling of skeletons and to learn a multi-modal skeleton representation by leveraging the relative positions of joints. However, adapting these models to the spatio-temporal graphs constructed based on facial landmarks present many challenges, including structural differences, higher dimensionality of nodes, fine-grained muscle movements, etc. As a result, these models cannot be directly applied to spatio-temporal graphs constructed from facial landmarks.

\section{Methodology}~\label{sec3}
In this section, we give a formal problem definition of driver fatigue detection and describe the general framework of our model LiteFat, as shown in Fig.~\ref{fig:fm}. LiteFat on the highest level comprises four components, namely \textit{facial landmark detection}, 
\textit{enhanced feature extraction}, \textit{multimodal feature fusion}, and \textit{STGL-based driver fatigue detection}. At first, we generate a sequence of 2-D coordinates along with its confidence scores to represent the driver's facial landmarks in each frame. Simultaneously, MobileNet-V3~\cite{howard2019searching} is applied to generate enhanced embeddings capturing driver's additional facial features. After these two steps, an informative feature matrix for the facial graph is generated by learning a multimodal fusion module to fuse the information of landmark positions and enhanced facial embeddings. The final step is to learn a spatio-temporal GNN for driver fatigue detection. In this component, the adjacency matrix of the graph is learned adaptively instead of being defined by rules or heuristics, which can avoid incorporating human bias into the spatial relationship among those facial landmarks. In the rest of this section, we introduce the four main components of LiteFat in detail. 
\subsection{Problem Definition}
In this paper, we use bold lower/uppercase case letters to represent vectors/matrices. The sets are denoted by calligraphic fonts (e.g., $\mathcal{V}$). Accordingly, the driver fatigue detection problem can be formally stated as a classification problem:

In each video clip composed of $T$ frames, each frame is represented by a graph $G=(\mathcal{V},\mathcal{E})$, where $\mathcal{V}$ refers to the node set and $\mathcal{E}$ represents the edge set. The adjacency matrix of a graph is represented by $\textbf{A}\in\textbf{R}^{N\times N}$, where $\textbf{A}_{ij}$ is 1 if $(v_i,v_j)\in E$ otherwise it is 0. At each timestamp $t$, the graph $G$ has its specific feature matrix $\textbf{X}^{(t)}\in \textbf{R}^{N\times D}$. Given a graph $G$ and the feature matrices of its historical $S$ frames, our problem is to learn a classifier $f$ to predict the label of each frame.

\subsection{Facial Landmark Detection}
Facial landmark detection algorithms aim to identify key points on the human face, such as the eyes, nose, mouth, and jawline~\cite{zhu2019robust}. These algorithms are essential for applications like facial recognition, emotion analysis, and driver fatigue detection~\cite{zhang2018combining,sukno20143}.

\begin{figure}[thpb]
      \centering
      \includegraphics[scale=0.35]{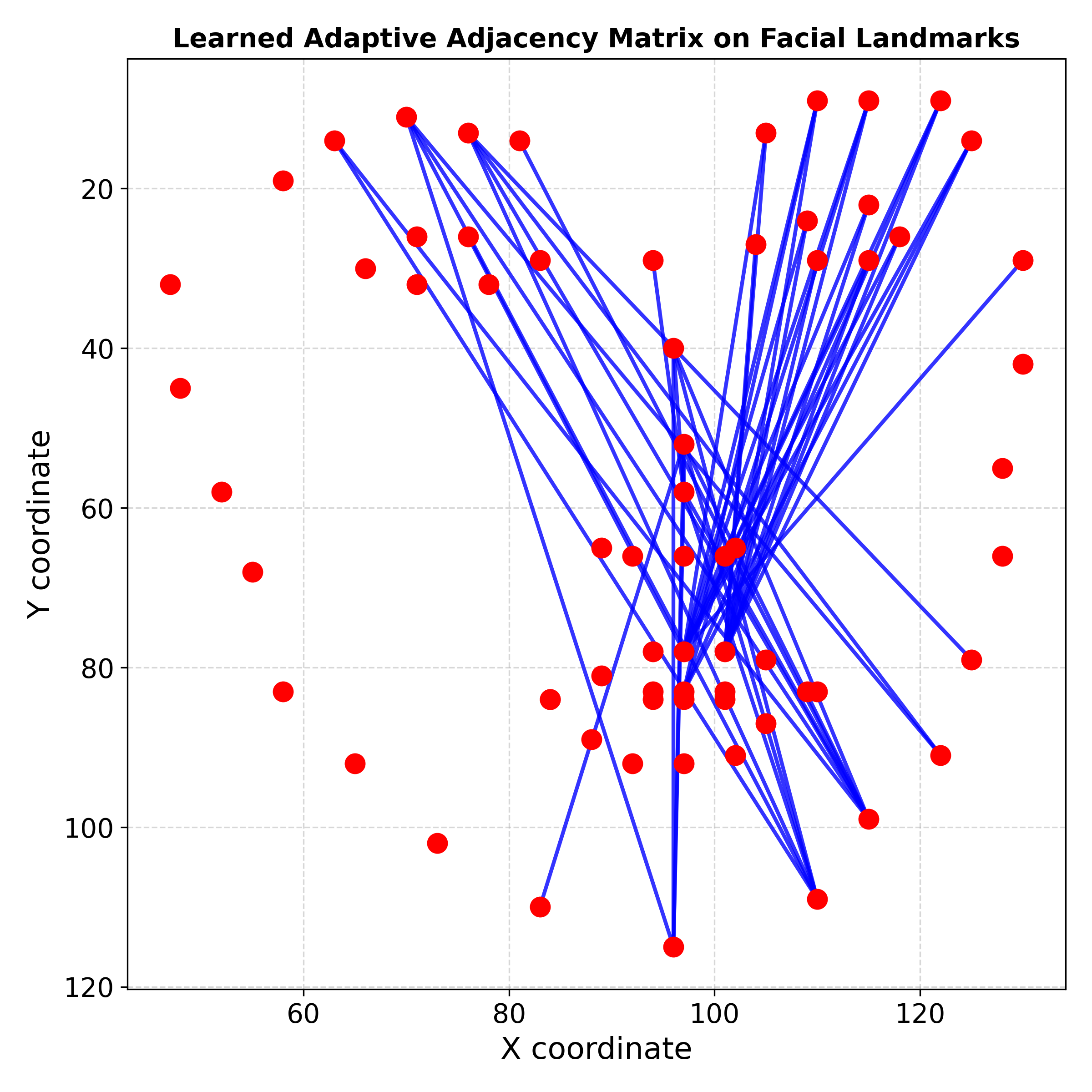}
      \caption{Visualization of the adaptive adjacency matrix on facial landmarks, where only the top 50 weighted edges are selected for clarity.}
      \label{fig:adj}
\end{figure}

As shown in Fig.~\ref{fig:adj}, there are totally 68 facial landmarks in each human face. For the facial landmark points, the algorithm~\cite{bulat2017far} represents each node as $(X,Y,C)$, where $(X,Y)$ denotes a 2-D coordinate in coordinate system and $C$ is the confidence score. Then, each frame in a video is represented by a $68\times 3$ matrix $\textbf{C}$. 

In practice, facial landmark detection may fail to capture key facial points when the driver looks down or turns their head away. To handle this situation, we set the values of every element in the coordinate matrix into 1, with the aim of ensuring this will not influence the consequent calculation for  feature matrix construction.
\subsection{Feature Matrix Construction}
Different from general network/graph data, there is no feature matrix as the input of STGL model. To enable the model to learn more information from driver face, environments, and driver behaviors, we propose to fuse multimodal features from facial landmarks and videos to create a more informative feature matrix.

\subsubsection{MobileNet for enhanced embedding}
Despite facial landmarks can detect structured geometric information about key facial points, features like skin tone, wrinkles, or redness that are also important for fatigue detection may be lose. To compensate for these lost information, we apply a lightweight model, MobileNet, to learn additional features and combine them with coordinates for feature matrix construction.

MobileNet is a family of lightweight deep learning models~\cite{sandler2018mobilenetv2,howard2019searching} designed for efficient computer vision tasks, particularly on mobile and embedded devices. It was introduced to provide a good balance between accuracy and computational efficiency. In our model, we choose MobileNetV3-Small~\cite{howard2019searching} to learn a $D$ dimensional vector $\textbf{d}$ for each frame to capture additional facial and contextual information.
\subsubsection{Multimodal fusion module}
The multimodal fusion module is a core component for our feature matrix construction. It fuses the embeddings of the facial landmark and MobileNet and outputs the final feature martix. The feature matrix can be calculated by:
\begin{equation}
    \textbf{X}=\textbf{C}\textbf{w}\textbf{d}^T ,
\end{equation}
where $\textbf{w}$ is the weight vector of multimodal fusion module.
\subsection{Spatial Temporal Graph Learning}
\subsubsection{Graph Convolution Layer}
Graph convolution is a feature aggregation operation where each node updates its representation by aggregating features from its neighbors~\cite{kipf2017semi}. Instead of using higher-order Chebyshev polynomials~\cite{defferrard2016convolutional}, GCN simplifies the convolution operation to a first-order approximation, making it efficient and easy to implement. Let $\tilde
{\textbf{A}} \in R^{N\times N}$, $\textbf{X} \in R^{N\times D}$, and $\textbf{Z} \in R^{N\times M}$ denote the normalized adjacency matrix, input feature matrix, and the output, respectively.


\textbf{Self-adaptive Adjacency Matrix:}
As shown in Fig.~\ref{fig:fm}, there is no internal links among facial landmarks. Considering that defining relationships based on rules may incorporate human bias or lose hidden spatial dependencies,
inspired by~\cite{wu2019graph}, we propose a self-adaptive adjacency matrix $\tilde
{\textbf{A}}_{adp}$ to learn by itself in an end-to-end manner through stochastic gradient descent. Another advantage of adaptive adjacency matrix is that it does not require any interruption from human experts. Specifically, we randomly initialize two node embedding dictionaries with learnable parameters $\textbf{E}_1, \textbf{E}_2 \in R^{N\times c}$, then the self-adaptive adjacency matrix is learned as:
\begin{equation}
    \tilde
{\textbf{A}}_{adp} \rm=softmax(ReLU(\textbf{E}_1 \textbf{E}_2^T)).
\end{equation}
Here, $\textbf{E}_1$ and $\textbf{E}_2$ denote the source and target node embedding. The spatial dependency weights between the source and target nodes could be derived by multiplying $\textbf{E}_1$ and $\textbf{E}_2$. The weak connections could be eliminated by the ReLU activation function, and the softmax function is used to normalize the self-adaptive adjacency matrix. After the process of adaptive adjacency matrix learning, our model could be learned as:
\begin{equation}
    \textbf{Z} = F(\textbf{X},\tilde
{\textbf{A}}_{adp})=\rm softmax(\tilde
{\textbf{A}}_{adp}ReLU(\tilde
{\textbf{A}}_{adp}\textbf{X}\textbf{W}^{(0)})\textbf{W}^{(1)}),
\end{equation}
where $\textbf{W}^{(0)}\in R^{D\times H}$ is a weight matrix from input layer to hidden layer with $H$ feature maps, and $\textbf{W}^{(1)} \in R^{D\times M}$ is the hidden-to-output weight matrix. The softmax function $\rm softmax(x_i)=\frac{1}{z}exp(x_i)$ with $\rm z=\sum_{i}exp(x_i)$ is applied row-wise.
\subsubsection{Temporal Convolution Layer}
\begin{figure}[thpb]
      \centering
     
      \includegraphics[scale=0.6]{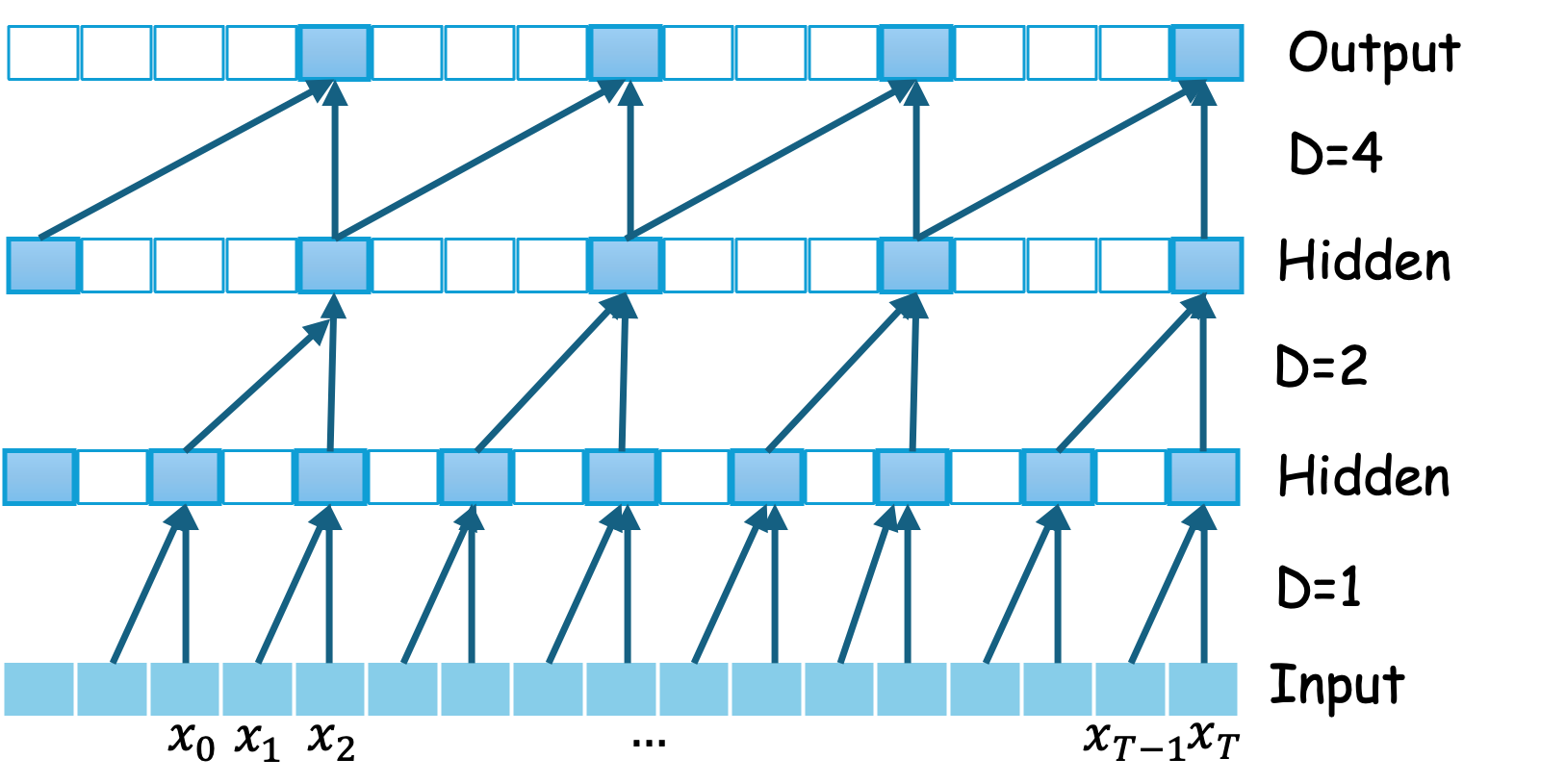}
      \caption{A dilated causal convolution with dilation factors d = 1, 2, 4 and filter size k = 2.}
      \label{fig:tcn}
   \end{figure}

Experiments have shown that Temporal Convolutional Networks (TCNs)~\cite{bai2018empirical} can outperform LSTMs and GRUs in sequence modeling, and convolutions provide better parallelism, stable gradients, and longer memory than RNNs. Therefore, TCNs have increasingly become an alternative to recurrent neural networks (RNNs) in recent years. As opposed to RNN-based approaches, TCNs use dilated causal convolutions to model long-range dependencies in a non-recursive manner, and the causal structure ensures that predictions only depend on past inputs (i.e., it does not ``leak" future information). Specifically, the dilated causal convolution, as shown in Fig.~\ref{fig:tcn}, processes inputs by moving across them while skipping values at fixed intervals. Therefore, we adopt the dilated causal convolution as our temporal convolution layer (TCN) to capture a node's temporal trends. Mathematically, for a 1-D sequence input $\textbf{x} \in R^n$ and a filter $f:\{0,...,k-1\} \to R$, the dilated causal convolution operation of $\textbf{x}$ with $f$ at step $t$ is represented as
\begin{equation}
    \textbf{x}\star f(t)=\sum_{s=0}^{k-1}f(s)\textbf{x}(t-d\times s),
\end{equation}
where the dilation rate $d$ determines the spacing between elements in the convolution kernel. Stacking dilated causal convolution layers with progressively increasing dilation factors allows the model's receptive field to expand exponentially. This enables dilated causal convolution networks to capture long-range dependencies using fewer layers, reducing computational cost.

\textbf{Gated TCN.}
Gating mechanisms play a crucial role in RNNs and have also proven effective in regulating information flow across layers in temporal convolution networks~\cite{dauphin2017language}. Given the input $\mathcal{X}\in R^{N\times D\times T}$, the Gated TCN is calculated as:
\begin{equation}
    \textbf{h} = g(\Theta_1 \star \mathcal{X} + \textbf{b})\odot \sigma(\Theta_2 \star \mathcal{X} + \textbf{c}),
\end{equation}
where $\Theta_1$, $\Theta_2$, $\textbf{b}$, and $\textbf{c}$ are model parameters, and $\odot$ refers to element-wise product. Here, we set the tangent hyperbolic function as the activation function $g(\cdot)$, and $\sigma(\cdot)$ is the sigmod function determining which information can be passed to the next layer.

\subsubsection{STGL Model}
A brief framework of our STGL model is shown in Fig.~\ref{fig:fm}. It is composed of stacked spatial-temporal layers and a final output layer. Each spatial-temporal layer integrates two GCN layers with a gated TCN, which consists of two parallel temporal convolution layers (TCN-a and TCN-b). By stacking multiple spatial-temporal layers, it effectively captures spatial dependencies across different temporal scales. Specifically, in the lower layers, the GCN processes short-term temporal information, while in the upper layers, it focuses on long-term temporal dependencies.

\textbf{Objective Function}: The inputs to a graph convolution layer in practice are three-dimension tensors with size [\textit{N}, \textit{D}, \textit{T}] where \textit{N} is the number of nodes, \textit{D} is the hidden dimension, and \textit{T} is the number of frames. We choose to use cross entropy as the training objective of LiteFat:
\begin{equation}
    L=-\frac{1}{T}\sum_{i=1}^{T}y_{i}\rm log(\hat{y}_{i}).
\end{equation}

\section{Experiments}~\label{sec4}
In this section, experiments are carried out on the YawDD dataset~\cite{abtahi2014yawdd} to evaluate the performance of LiteFat in comparison with baselines. The datasets, introduction of baseline methods, evaluation metrics used in this experiment, implementation and training details, and experimental result analysis are described in the following subsections, respectively.
\subsection{Dataset}
YawDD~\cite{abtahi2014yawdd} is a comprehensive collection of video data aimed at facilitating research in driver fatigue detection. It consists of diverse video sequences capturing normal, yawning and talking behaviors, featuring varied lighting conditions, two angles, and different participant demographics to simulate real-world driving scenarios. There are eight scenarios in the dataset, according to the male/female with/without glasses filmed from the mirror/dash of the car. Note that there is another dataset NTHU-DDD~\cite{weng2017driver}, which is not publicly available.

Since LiteFat feeds on videos instead of frames,
all videos in YawDD are divided into video clips containing only one action, and we cut all YawDD data into video clips within 5 seconds. We relabeled all video clips, and the labels of all frames are the same, which is classified as normal (N), talking (T), and yawning (Y). 
Note that the videos labeled as talking also include instances where the driver turns their head and their eyes are not looking straight ahead. We treat these videos as an abnormal state of distracted driving. Therefore, the comparative experiment of yawning detection is a binary classification task (i.e., yawning and normal), while all ablation experiments are based on a three-class classification, namely normal, yawning, and talking. This means that our model can not only detect driver fatigue but also identify when the driver is distracted. In actual system deployment, different levels of warnings can be given based on the type of abnormality.
\subsection{Baseline Methods}
\begin{itemize}
    \item \textbf{2s-STGCN}~\cite{bai2021two} is a two-stream architecture, which uses two distinct processing pathways to capture spatial features of facial landmarks and temporal features that are crucial for capturing the ongoing patterns or sequences of behavior.
    \item \textbf{JHPFA-Net}~\cite{lu2023jhpfa} is a head pose and face action fusion module to combine head pose estimation and facial action recognition jointly.
    \item \textbf{2s-IsoSSL-MoCo}~\cite{mou2021isotropic} is an attention-based multimodal fusion model to fuse features from the eye, mouth, and optical ﬂow of the head for driver drowsiness detection.
\end{itemize}
In this paper we directly referenced the results reported in the paper presenting each baseline method using the same dataset. Furthermore, we ran efficiency experiments by reproducing these methods described in those papers.

\subsection{Evaluation Metrics}
For comparison experiments, we select accuracy, precision, recall, F1-score, and Area Under the Curve (AUC) as the evaluation metrics. For efficiency verification, we compare the number of parameters, the forward and backward pass times during each batch, the number of samples it can process per second (throughput), and memory usage of the models. 

\subsection{Implementation Details}
The hardware experimental environment is as follows. The chip is an Apple M3 Pro with 12 cores (6 performance and 6 efficiency). The memory size is 18 GB. The software experimental environment is as follows. The operation system is MacOS 15.0.1 (24A348), and our method is coded in Python with PyTorch 2.6.0. 

The proposed fatigue detection model was trained using a preprocessed dataset in which each sample consists of a video represented by 16 frames. The model was trained for a maximum of 100 epochs, with an early stopping strategy implemented to terminate training if the training loss did not improve for three consecutive epochs. For optimization, we used the Adam optimizer with an initial learning rate of $1\times10^{-4}$. 

\subsection{Experimental Analysis}
\subsubsection{Comparison of LiteFat with State-of-the-Art Methods}
Table~\ref{table_perf} presents a performance comparison of various yawning detection methods on the YawDD dataset, evaluating accuracy, precision, recall, and F1-score. LiteFat achieves the best overall performance, attaining high accuracy, recall, and F1-score, making it the most effective model. 2s-IsoSSL-MoCo follows closely with an accuracy of 0.987 and an F1-score of 0.934, though its precision and recall values are not provided. JHPFA-Net demonstrates the second-highest precision but has a lower recall. 2s-STGCN and spatial stream models perform reasonably well, with accuracies of 0.934 and 0.930, respectively, but have lower F1-scores. The temporal stream method lags behind, with the lowest precision and F1-score. Overall, LiteFat outperforms all other methods, making it the most robust and accurate model for yawning detection in this study.

\begin{figure}[thpb]
      \centering
     
      \includegraphics[scale=0.5]{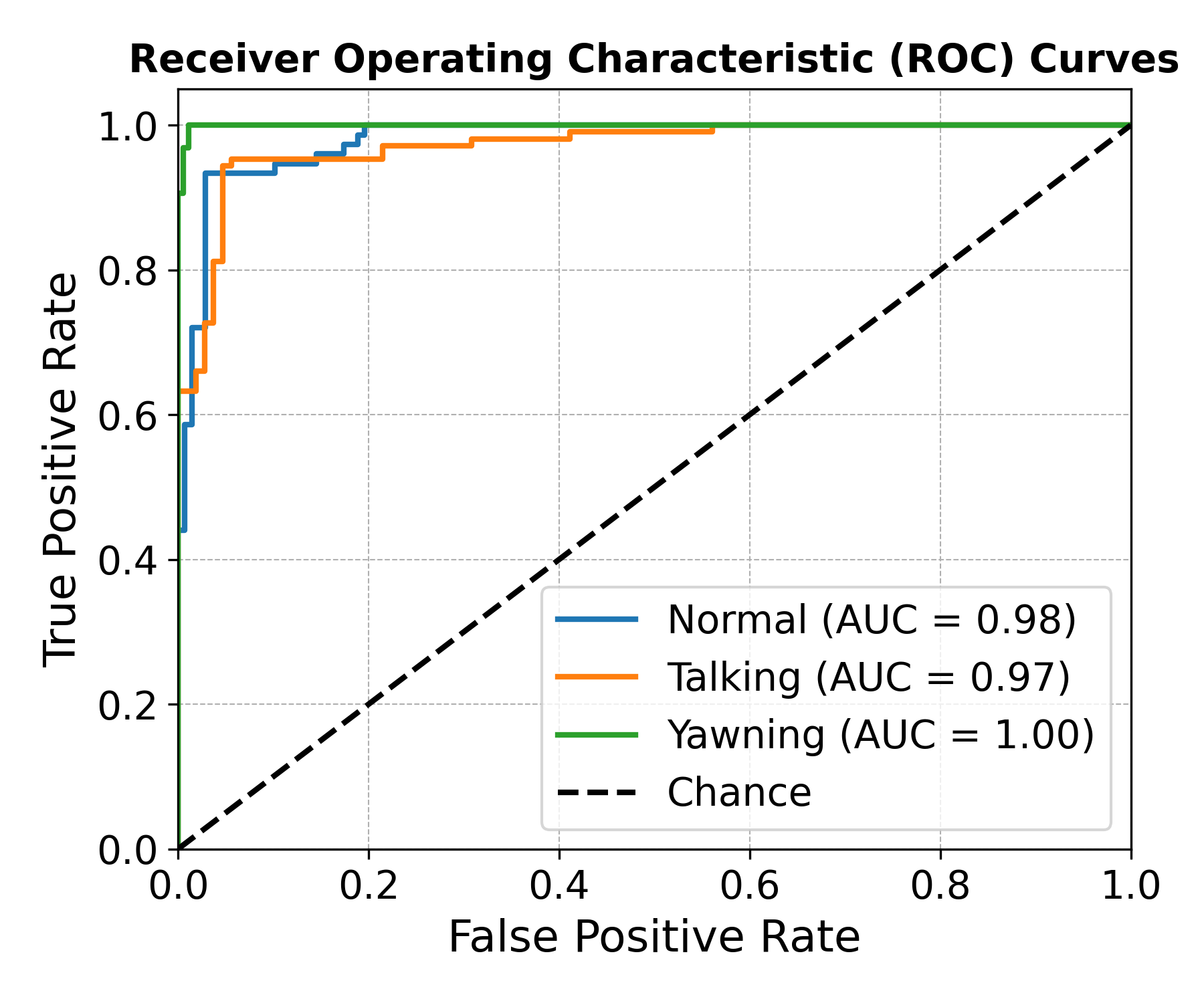}
      \caption{AUC of our model in the three-class classification task, namely normal, talking, and yawning. A higher AUC value indicates better model performance.}
      \label{fig:rocs}
   \end{figure}

Figure.~\ref{fig:rocs} presents the Receiver Operating Characteristic (ROC) curves for a three-class classification task, distinguishing between normal, talking, and yawning states. The AUC values for each class are provided, indicating the model's discriminative performance. The ``Yawning" class achieves a perfect AUC of 1.00, suggesting flawless classification. The ``Normal" and ``Talking" classes also exhibit strong performance, with AUC values of 0.98 and 0.97, respectively. The ROC curves remain close to the upper-left corner, indicating high true positive rates with low false positive rates. The dashed diagonal line represents random chance classification, and the model's curves significantly surpass this baseline, demonstrating its strong predictive ability. Overall, the high AUC values suggest that our model performs exceptionally well in differentiating between the three classes.
\begin{table}[h]
\caption{Performance comparison of various methods for yawning detection on the YawDD dataset. The \textbf{bold} number indicates the best result and the underlined number indicates the second best result.}
\label{table_perf}
\begin{center}
\begin{tabular}{ccccc}
\hline
\textbf{Method} & Acc.&Pre.&Recall&F1-score\\
\hline

\textbf{spatial stream}~\cite{bai2021two} &0.930&0.910&\underline{0.960}&0.930\\
\textbf{temporal stream}~\cite{bai2021two} &0.920&0.824&0.933&0.875\\
\textbf{2s-STGCN}~\cite{bai2021two} &0.934&0.855&0.940&0.895\\
\textbf{JHPFA-Net}~\cite{lu2023jhpfa} &0.964&\underline{0.977}&0.903&0.939\\
\textbf{2s-IsoSSL-MoCo}~\cite{mou2021isotropic} &\underline{0.987}&-&-&\underline{0.983}\\
\textbf{LiteFat (ours)} &\textbf{0.995}
&\textbf{1.00}&\textbf{0.969}&\textbf{0.984}\\
\hline
\end{tabular}
\end{center}
\end{table}

\begin{table}[h]\large
\caption{Efficiency comparison of two fatigue detection methods on YawDD dataset. \#para. refers to the parameter count used in the model, forw. and back. show the pass time (sec.) of each batch in forward and backward, thr. means the throughput (samples/sec) of the model, and mem. denotes the memory usage (MB) of the models.}
\label{table_efficiency}
\begin{center}
\resizebox{8.5cm}{!}{
\begin{tabular}{cccccc}
\hline
\textbf{Method} &\#para.& forw. &back.&thr.& mem.\\
\hline
\textbf{2s-STGCN}~\cite{bai2021two} &9.97E5&1.52&0.30&5.26&1367.44\\
\textbf{JHPFA-Net}~\cite{lu2023jhpfa} &2.26E8&10.35&14.77&0.77&2201.03\\
\textbf{2s-IsoSSL-MoCo}~\cite{mou2021isotropic} &2.48E7&1.89&3.11&4.23&3433.48\\
\textbf{LiteFat (ours)} &1.32E6&1.39&2.16&5.75&1905.44\\
\hline
\end{tabular}
}
\end{center}
\end{table}


\subsubsection{Efficiency Analysis}
Table~\ref{table_efficiency} presents the efficiency comparison of between three SOTA fatigue detection baselines and LiteFat on the YawDD dataset, focusing on their parameter count, forward and backward pass times, throughput, and memory usage. In this experiment, a 2-second dummy video is used as input to compare the parameters of different models. As shown in the table, with only 1.32E6 parameters, LiteFat is significantly more lightweight than JHPFA-Net and 2s-IsoSSL-MoCo. LiteFat
achieves a forward pass time of 1.39 seconds, which is faster than both 2s-STGCN and 2s-IsoSSL-MoCo, and significantly more efficient than JHPFA-Net, which records a forward pass time of 10.35 seconds. In terms of backward pass time, LiteFat records 2.16 seconds, which is approximately 6.8 times faster than JHPFA-Net's 14.77 seconds, demonstrating its efficiency in training scenarios. LiteFat also leads in throughput, processing 5.75 samples per second, which surpasses all other methods, indicating its capability for real-time applications. Additionally, LiteFat's memory usage is 1905.44 MB which is slightly higher than 2s-STGCN but significantly lower than 2s-IsoSSL-MoCo and JHPFA-Net, highlighting its resource efficiency. 


\subsubsection{Ablation Study}

\textbf{Importance of STGL and MobileNet.} Table~\ref{table_ablation} presents an ablation study evaluating the impact of STGL and MobileNet on model performance across five metrics: AUC, Accuracy, Precision, Recall, and F1-score. The results indicate that using MobileNet alone leads to a significant performance boost compared to using STGL alone, with an accuracy of 0.92 versus 0.83. Similarly, AUC value, precision, recall, and F1-score are all higher for MobileNet alone compared to STGL alone. However, the best performance is achieved when both STGL and MobileNet are combined, resulting in the highest AUC value (0.998), accuracy (0.977), precision (0.974), recall (0.984), and F1-score (0.978). This suggests that while MobileNet is a strong feature extractor, integrating STGL further enhances model performance, demonstrating their complementary nature.

\begin{table}[h]
\caption{Ablation studies of the model evaluating the impact of STGL and MobileNet on fatigue \& distraction detection across four metrics.}
\label{table_ablation}
\begin{center}
\begin{tabular}{ccccccccc}
\hline
STGL & MobileNet&AUC&Acc.&Pre.&Recall&F1-score\\
\hline
 \checkmark&\usym{2613}&0.942&0.826&0.873&0.830&0.844\\
 \usym{2613}&\checkmark&0.975&0.920&0.940&0.910&0.922\\
 
 \checkmark&\checkmark&\textbf{0.998}&\textbf{0.977}&\textbf{0.974}&\textbf{0.984}&\textbf{0.978}\\

\hline
\end{tabular}
\end{center}
\end{table}

\textbf{Importance of Spatial GCN and TCN.} 
Table~\ref{table_stablation} presents the ablation study results for the spatial-temporal module, evaluating the impact of removing either TCN or GCN on model performance. STGL (which is used in our LiteFat model) achieves the highest performance across all metrics, indicating its effectiveness in capturing both spatial and temporal dependencies. When TCN is removed (w/o TCN), performance declines slightly, with AUC dropping to 0.981, accuracy to 0.892, and F1-score to 0.904, suggesting that temporal features contribute to model effectiveness but are not solely responsible for strong performance. However, removing GCN (w/o GCN) results in a more significant drop, with AUC decreasing to 0.974, accuracy to 0.864, and F1-score to 0.871, highlighting the critical role of spatial feature extraction in the model. These results demonstrate that both spatial and temporal components are essential, but spatial modeling (GCN) appears to have a slightly greater impact on overall performance.
\begin{table}[h]
\caption{Spatial-temporal module ablation experimental results.}
\label{table_stablation}
\begin{center}
\begin{tabular}{cccccc}
\hline
Method & AUC&Acc.&Pre.&Recall&F1-score\\
\hline
w/o TCN&0.981&0.892&0.917&0.897&0.904\\
w/o GCN &0.974&0.864&0.895&0.860&0.871\\
STGL&\textbf{0.998}&\textbf{0.977}&\textbf{0.974}&\textbf{0.984}&\textbf{0.978}\\

\hline
\end{tabular}
\end{center}
\end{table}

\textbf{Importance of Adaptive Adjacency Matrix.} 
To verify the effectiveness of the adaptive adjacency matrix, this section uses five different adjacency matrix configurations to conduct experiments with STGL. The adjacency matrix we select are: (1) Normalized Laplacian Matrix;
(2) Scaled Laplacian Matrix;
(3) Symmetrically Normalized Adjacency Matrix;
(4) Asymmetrically Normalized Adjacency Matrix;
and (5) Identity Matrix.

Table~\ref{table_adjacencyablation} presents the performance comparison of five methods using various evaluation metrics. Among these, the adaptive adjacency matrix outperforms all others, achieving the highest values of all metrics, indicating superior overall performance. The experimental results show that, compared to fixed-format adjacency matrices, the adjacency matrix learned adaptively by the model can more comprehensively capture facial structural information. Additionally, as time progresses, the structure captured by the model also changes.
\\

\begin{table}[h]
\caption{Adjacency matrix comparison}
\label{table_adjacencyablation}
\begin{center}
\begin{tabular}{cccccc}
\hline
\textbf{Method} & AUC&Acc.&Pre.&Recall&F1-score\\
\hline
\textbf{Norm. lap.}&0.973&0.887&0.910&0.900&0.903\\
\textbf{Scal. lap.}&0.918&0.704&0.793&0.693&0.699\\
\textbf{Sysm. norm.}&0.971&0.892&0.917&0.897&0.901\\
\textbf{Asysm. norm.}&0.969&0.892&0.920&0.86&0.887\\
\textbf{Identity} &0.957&0.840&0.830&0.827&0.826\\
\textbf{Adaptive}&\textbf{0.998}&\textbf{0.977}&\textbf{0.974}&\textbf{0.984}&\textbf{0.978}\\
\hline
\end{tabular}
\end{center}
\end{table}

\section{Conclusion}~\label{sec5}
In this paper, we have proposed LiteFat, a lightweight and efficient graph learning model for driver fatigue detection in embedded robotic systems. By integrating facial landmark-based features with embeddings learned from MobileNet, our approach effectively captures critical fatigue-related patterns while maintaining computational efficiency. The proposed model balances accuracy and speed, making it well-suited for real-time applications in resource-constrained environments, such as embedded and robotic devices in intelligent vehicles. Experimental results demonstrate that LiteFat achieves competitive performance compared to state-of-the-art methods while significantly reducing model complexity. Future work will implement the model upon  embedded robotic devices and evaluate its performance with more real-world data. Moreover, building on the current model, we will design additional modules to handle other modalities and integrate them into the existing architecture, enabling extension to multimodal datasets and further improving detection performance.




\bibliographystyle{IEEEtran}
\bibliography{IEEEabrv}

\begin{thebibliography}{10}
\providecommand{\url}[1]{#1}
\csname url@samestyle\endcsname
\providecommand{\newblock}{\relax}
\providecommand{\bibinfo}[2]{#2}
\providecommand{\BIBentrySTDinterwordspacing}{\spaceskip=0pt\relax}
\providecommand{\BIBentryALTinterwordstretchfactor}{4}
\providecommand{\BIBentryALTinterwordspacing}{\spaceskip=\fontdimen2\font plus
\BIBentryALTinterwordstretchfactor\fontdimen3\font minus \fontdimen4\font\relax}
\providecommand{\BIBforeignlanguage}[2]{{%
\expandafter\ifx\csname l@#1\endcsname\relax
\typeout{** WARNING: IEEEtran.bst: No hyphenation pattern has been}%
\typeout{** loaded for the language `#1'. Using the pattern for}%
\typeout{** the default language instead.}%
\else
\language=\csname l@#1\endcsname
\fi
#2}}
\providecommand{\BIBdecl}{\relax}
\BIBdecl

\bibitem{fu2024survey}
B.~Fu, F.~Boutros, C.-T. Lin, and N.~Damer, ``A survey on drowsiness detection--modern applications and methods,'' \emph{IEEE Transactions on Intelligent Vehicles}, 2024.

\bibitem{nvemcova2020multimodal}
A.~N{\v{e}}mcov{\'a}, V.~Svozilov{\'a}, K.~Bucsuh{\'a}zy, R.~Sm{\'\i}{\v{s}}ek, M.~M{\'e}zl, B.~Hesko, M.~Bel{\'a}k, M.~Bil{\'\i}k, P.~Maxera, M.~Seitl \emph{et~al.}, ``Multimodal features for detection of driver stress and fatigue,'' \emph{IEEE Transactions on Intelligent Transportation Systems}, vol.~22, no.~6, pp. 3214--3233, 2020.

\bibitem{liu2022review}
F.~Liu, D.~Chen, J.~Zhou, and F.~Xu, ``A review of driver fatigue detection and its advances on the use of rgb-d camera and deep learning,'' \emph{Engineering Applications of Artificial Intelligence}, vol. 116, p. 105399, 2022.

\bibitem{lu2022detecting}
K.~Lu, A.~S. Dahlman, J.~Karlsson, and S.~Candefjord, ``Detecting driver fatigue using heart rate variability: A systematic review,'' \emph{Accident Analysis \& Prevention}, vol. 178, p. 106830, 2022.

\bibitem{hu2024eeg}
F.~Hu, L.~Zhang, X.~Yang, and W.-A. Zhang, ``Eeg-based driver fatigue detection using spatio-temporal fusion network with brain region partitioning strategy,'' \emph{IEEE Transactions on Intelligent Transportation Systems}, 2024.

\bibitem{peng2024multi}
Y.~Peng, H.~Deng, G.~Xiang, X.~Wu, X.~Yu, Y.~Li, and T.~Yu, ``A multi-source fusion approach for driver fatigue detection using physiological signals and facial image,'' \emph{IEEE Transactions on Intelligent Transportation Systems}, 2024.

\bibitem{subasi2022eeg}
A.~Subasi, A.~Saikia, K.~Bagedo, A.~Singh, and A.~Hazarika, ``Eeg-based driver fatigue detection using fawt and multiboosting approaches,'' \emph{IEEE Transactions on Industrial Informatics}, vol.~18, no.~10, pp. 6602--6609, 2022.

\bibitem{ansari2021driver}
S.~Ansari, F.~Naghdy, H.~Du, and Y.~N. Pahnwar, ``Driver mental fatigue detection based on head posture using new modified relu-bilstm deep neural network,'' \emph{IEEE Transactions on Intelligent Transportation Systems}, vol.~23, no.~8, pp. 10\,957--10\,969, 2021.

\bibitem{ren2021deep}
J.~Ren, F.~Xia, Y.~Liu, and I.~Lee, ``Deep video anomaly detection: Opportunities and challenges,'' in \emph{2021 international conference on data mining workshops (ICDMW)}.\hskip 1em plus 0.5em minus 0.4em\relax IEEE, 2021, pp. 959--966.

\bibitem{sikander2018driver}
G.~Sikander and S.~Anwar, ``Driver fatigue detection systems: A review,'' \emph{IEEE Transactions on Intelligent Transportation Systems}, vol.~20, no.~6, pp. 2339--2352, 2018.

\bibitem{zhang2022systematic}
Z.~Zhang, H.~Ning, and F.~Zhou, ``A systematic survey of driving fatigue monitoring,'' \emph{IEEE transactions on intelligent transportation systems}, vol.~23, no.~11, pp. 19\,999--20\,020, 2022.

\bibitem{kaur2024deepfake}
A.~Kaur, A.~Noori~Hoshyar, V.~Saikrishna, S.~Firmin, and F.~Xia, ``Deepfake video detection: challenges and opportunities,'' \emph{Artificial Intelligence Review}, vol.~57, no.~6, pp. 1--47, 2024.

\bibitem{shahzad2024decade}
M.~Z. Shahzad, M.~A. Hanif, and M.~Shafique, ``Decade: Towards designing efficient-yet-accurate distance estimation modules for collision avoidance in mobile advanced driver assistance systems,'' in \emph{2024 IEEE/RSJ International Conference on Intelligent Robots and Systems (IROS)}.\hskip 1em plus 0.5em minus 0.4em\relax IEEE, 2024, pp. 334--340.

\bibitem{li2024gmmap}
P.~Z.~X. Li, S.~Karaman, and V.~Sze, ``Gmmap: Memory-efficient continuous occupancy map using gaussian mixture model,'' \emph{IEEE Transactions on Robotics}, vol.~40, pp. 1339--1355, 2024.

\bibitem{yamada2024twist}
J.~Yamada, M.~Rigter, J.~Collins, and I.~Posner, ``Twist: Teacher-student world model distillation for efficient sim-to-real transfer,'' in \emph{2024 IEEE International Conference on Robotics and Automation (ICRA)}.\hskip 1em plus 0.5em minus 0.4em\relax IEEE, 2024, pp. 9190--9196.

\bibitem{ren2023graph}
J.~Ren, F.~Xia, I.~Lee, A.~Noori~Hoshyar, and C.~Aggarwal, ``Graph learning for anomaly analytics: Algorithms, applications, and challenges,'' \emph{ACM Transactions on Intelligent Systems and Technology}, vol.~14, no.~2, pp. 1--29, 2023.

\bibitem{xu2025fairness}
Z.~Xu, S.~Kandanaarachchi, C.~S. Ong, and E.~Ntoutsi, ``Fairness evaluation with item response theory,'' in \emph{Proceedings of the ACM on Web Conference 2025}, 2025, pp. 2276--2288.

\bibitem{lin2025early}
C.~Lin, X.~Zhu, R.~Wang, W.~Zhou, N.~Li, and Y.~Xie, ``Early driver fatigue detection system: A cost-effective and wearable approach utilizing embedded machine learning,'' \emph{Vehicles}, vol.~7, no.~1, p.~3, 2025.

\bibitem{wang2021survey}
J.~Wang, W.~Chai, A.~Venkatachalapathy, K.~L. Tan, A.~Haghighat, S.~Velipasalar, Y.~Adu-Gyamfi, and A.~Sharma, ``A survey on driver behavior analysis from in-vehicle cameras,'' \emph{IEEE Transactions on Intelligent Transportation Systems}, vol.~23, no.~8, pp. 10\,186--10\,209, 2021.

\bibitem{bai2021two}
J.~Bai, W.~Yu, Z.~Xiao, V.~Havyarimana, A.~C. Regan, H.~Jiang, and L.~Jiao, ``Two-stream spatial--temporal graph convolutional networks for driver drowsiness detection,'' \emph{IEEE Transactions on Cybernetics}, vol.~52, no.~12, pp. 13\,821--13\,833, 2021.

\bibitem{yang2024video}
L.~Yang, H.~Yang, H.~Wei, Z.~Hu, and C.~Lv, ``Video-based driver drowsiness detection with optimised utilization of key facial features,'' \emph{IEEE Transactions on Intelligent Transportation Systems}, 2024.

\bibitem{lu2023jhpfa}
Y.~Lu, C.~Liu, F.~Chang, H.~Liu, and H.~Huan, ``Jhpfa-net: Joint head pose and facial action network for driver yawning detection across arbitrary poses in videos,'' \emph{IEEE Transactions on Intelligent Transportation Systems}, vol.~24, no.~11, pp. 11\,850--11\,863, 2023.

\bibitem{huang2024self}
Y.~Huang, C.~Liu, F.~Chang, and Y.~Lu, ``Self-supervised multi-granularity graph attention network for vision-based driver fatigue detection,'' \emph{IEEE Transactions on Emerging Topics in Computational Intelligence}, 2024.

\bibitem{mou2021isotropic}
L.~Mou, C.~Zhou, P.~Xie, P.~Zhao, R.~Jain, W.~Gao, and B.~Yin, ``Isotropic self-supervised learning for driver drowsiness detection with attention-based multimodal fusion,'' \emph{IEEE Transactions on Multimedia}, vol.~25, pp. 529--542, 2021.

\bibitem{phan2023driver}
A.-C. Phan, T.-N. Trieu, and T.-C. Phan, ``Driver drowsiness detection and smart alerting using deep learning and iot,'' \emph{Internet of Things}, vol.~22, p. 100705, 2023.

\bibitem{ahmed2021intelligent}
M.~Ahmed, S.~Masood, M.~Ahmad, and A.~A. Abd El-Latif, ``Intelligent driver drowsiness detection for traffic safety based on multi cnn deep model and facial subsampling,'' \emph{IEEE transactions on intelligent transportation systems}, vol.~23, no.~10, pp. 19\,743--19\,752, 2021.

\bibitem{sandler2018mobilenetv2}
M.~Sandler, A.~Howard, M.~Zhu, A.~Zhmoginov, and L.-C. Chen, ``Mobilenetv2: Inverted residuals and linear bottlenecks,'' in \emph{Proceedings of the IEEE conference on computer vision and pattern recognition}, 2018, pp. 4510--4520.

\bibitem{feng2022comparative}
L.~Feng, Y.~Zhao, W.~Zhao, and J.~Tang, ``A comparative review of graph convolutional networks for human skeleton-based action recognition,'' \emph{Artificial Intelligence Review}, pp. 1--31, 2022.

\bibitem{cheng2020skeleton}
K.~Cheng, Y.~Zhang, X.~He, W.~Chen, J.~Cheng, and H.~Lu, ``Skeleton-based action recognition with shift graph convolutional network,'' in \emph{Proceedings of the IEEE/CVF conference on computer vision and pattern recognition}, 2020, pp. 183--192.

\bibitem{zhang2020semantics}
P.~Zhang, C.~Lan, W.~Zeng, J.~Xing, J.~Xue, and N.~Zheng, ``Semantics-guided neural networks for efficient skeleton-based human action recognition,'' in \emph{proceedings of the IEEE/CVF conference on computer vision and pattern recognition}, 2020, pp. 1112--1121.

\bibitem{liu2020disentangling}
Z.~Liu, H.~Zhang, Z.~Chen, Z.~Wang, and W.~Ouyang, ``Disentangling and unifying graph convolutions for skeleton-based action recognition,'' in \emph{Proceedings of the IEEE/CVF conference on computer vision and pattern recognition}, 2020, pp. 143--152.

\bibitem{chen2021channel}
Y.~Chen, Z.~Zhang, C.~Yuan, B.~Li, Y.~Deng, and W.~Hu, ``Channel-wise topology refinement graph convolution for skeleton-based action recognition,'' in \emph{Proceedings of the IEEE/CVF international conference on computer vision}, 2021, pp. 13\,359--13\,368.

\bibitem{chi2022infogcn}
H.-g. Chi, M.~H. Ha, S.~Chi, S.~W. Lee, Q.~Huang, and K.~Ramani, ``Infogcn: Representation learning for human skeleton-based action recognition,'' in \emph{Proceedings of the IEEE/CVF conference on computer vision and pattern recognition}, 2022, pp. 20\,186--20\,196.

\bibitem{howard2019searching}
A.~Howard, M.~Sandler, G.~Chu, L.-C. Chen, B.~Chen, M.~Tan, W.~Wang, Y.~Zhu, R.~Pang, V.~Vasudevan \emph{et~al.}, ``Searching for mobilenetv3,'' in \emph{Proceedings of the IEEE/CVF international conference on computer vision}, 2019, pp. 1314--1324.

\bibitem{zhu2019robust}
M.~Zhu, D.~Shi, M.~Zheng, and M.~Sadiq, ``Robust facial landmark detection via occlusion-adaptive deep networks,'' in \emph{Proceedings of the IEEE/CVF conference on computer vision and pattern recognition}, 2019, pp. 3486--3496.

\bibitem{zhang2018combining}
H.~Zhang, Q.~Li, Z.~Sun, and Y.~Liu, ``Combining data-driven and model-driven methods for robust facial landmark detection,'' \emph{IEEE Transactions on Information Forensics and Security}, vol.~13, no.~10, pp. 2409--2422, 2018.

\bibitem{sukno20143}
F.~M. Sukno, J.~L. Waddington, and P.~F. Whelan, ``3-d facial landmark localization with asymmetry patterns and shape regression from incomplete local features,'' \emph{IEEE transactions on cybernetics}, vol.~45, no.~9, pp. 1717--1730, 2014.

\bibitem{bulat2017far}
A.~Bulat and G.~Tzimiropoulos, ``How far are we from solving the 2d \& 3d face alignment problem? (and a dataset of 230,000 3d facial landmarks),'' in \emph{International Conference on Computer Vision}, 2017.

\bibitem{kipf2017semi}
T.~N. Kipf and M.~Welling, ``Semi-supervised classification with graph convolutional networks,'' in \emph{International Conference on Learning Representations}, 2017.

\bibitem{defferrard2016convolutional}
M.~Defferrard, X.~Bresson, and P.~Vandergheynst, ``Convolutional neural networks on graphs with fast localized spectral filtering,'' \emph{Advances in neural information processing systems}, vol.~29, 2016.

\bibitem{wu2019graph}
Z.~Wu, S.~Pan, G.~Long, J.~Jiang, and C.~Zhang, ``Graph wavenet for deep spatial-temporal graph modeling,'' in \emph{Proceedings of the 28th International Joint Conference on Artificial Intelligence}, 2019, pp. 1907--1913.

\bibitem{bai2018empirical}
S.~Bai, J.~Z. Kolter, and V.~Koltun, ``An empirical evaluation of generic convolutional and recurrent networks for sequence modeling,'' in \emph{ICLR}, 2018.

\bibitem{dauphin2017language}
Y.~N. Dauphin, A.~Fan, M.~Auli, and D.~Grangier, ``Language modeling with gated convolutional networks,'' in \emph{International conference on machine learning}.\hskip 1em plus 0.5em minus 0.4em\relax PMLR, 2017, pp. 933--941.

\bibitem{abtahi2014yawdd}
S.~Abtahi, M.~Omidyeganeh, S.~Shirmohammadi, and B.~Hariri, ``Yawdd: A yawning detection dataset,'' in \emph{Proceedings of the 5th ACM multimedia systems conference}, 2014, pp. 24--28.

\bibitem{weng2017driver}
C.-H. Weng, Y.-H. Lai, and S.-H. Lai, ``Driver drowsiness detection via a hierarchical temporal deep belief network,'' in \emph{Computer Vision--ACCV 2016 Workshops: ACCV 2016 International Workshops, Taipei, Taiwan, November 20-24, 2016, Revised Selected Papers, Part III 13}.\hskip 1em plus 0.5em minus 0.4em\relax Springer, 2017, pp. 117--133.

\end{thebibliography}

\end{document}